\documentclass[letterpaper, 10 pt, conference]{ieeeconf}  

\IEEEoverridecommandlockouts                              

\usepackage{graphics} 
\usepackage{epsfig} 
\usepackage{hyperref}

\usepackage{soul} 
\usepackage{xcolor} 
\usepackage{pgfplots}
\usepackage{pgfplots}
\usepackage{comment}
\pgfplotsset{compat=1.8}
\usepgfplotslibrary{statistics}

\title{\LARGE \bf Hybrid Autonomy Framework for a Future Mars Science Helicopter}

\author{Luca Di Pierno$^{1,3}$, Robert Hewitt$^{2}$, Stephan Weiss$^{3}$ and Roland Brockers$^{2,3}$
\thanks{$^{1}$ ETH Zurich, Switzerland {\tt\small diluca@ethz.ch}}%
\thanks{$^{2}$ Jet Propulsion Laboratory / California Institute of Technology, Pasadena, CA, USA. {\tt\small \{robert.a.hewitt, roland.brockers\}@jpl.nasa.gov}}%
\thanks{$^{3}$ Control of Networked Systems Group, University of Klagenfurt, Austria. {\tt\small \{luca.dipierno, stephan.weiss, roland.brockers\}@aau.at}}%
}

\begin{document}
\maketitle
\thispagestyle{empty}
\pagestyle{empty}

\begin{abstract}

Autonomous aerial vehicles, such as NASA’s Ingenuity, enable rapid planetary surface exploration beyond the reach of ground-based robots. Thus, NASA is studying a \textit{Mars Science Helicopter (MSH)}, an advanced concept capable of performing long-range science missions and autonomously navigating challenging Martian terrain. Given significant Earth-Mars communication delays and mission complexity, an advanced autonomy framework is required to ensure safe and efficient operation by continuously adapting behavior based on mission objectives and real-time conditions, without human intervention. This study presents a deterministic high-level control framework for aerial exploration, integrating a Finite State Machine (FSM) with Behavior Trees (BTs) to achieve a scalable, robust, and computationally efficient autonomy solution for critical scenarios like deep space exploration. In this paper we outline key capabilities of a possible MSH and detail the FSM-BT hybrid autonomy framework which orchestrates them to achieve the desired objectives. Monte Carlo simulations and real field tests validate the framework, demonstrating its robustness and adaptability to both discrete events and real-time system feedback. These inputs trigger state transitions or dynamically adjust behavior execution, enabling reactive and context-aware responses. The framework is middleware-agnostic, supporting integration with systems like F-Prime and extending beyond aerial robotics.

\end{abstract}

\section{Introduction}

\begin{figure}[t]
    \centering
    \includegraphics[width=\columnwidth]{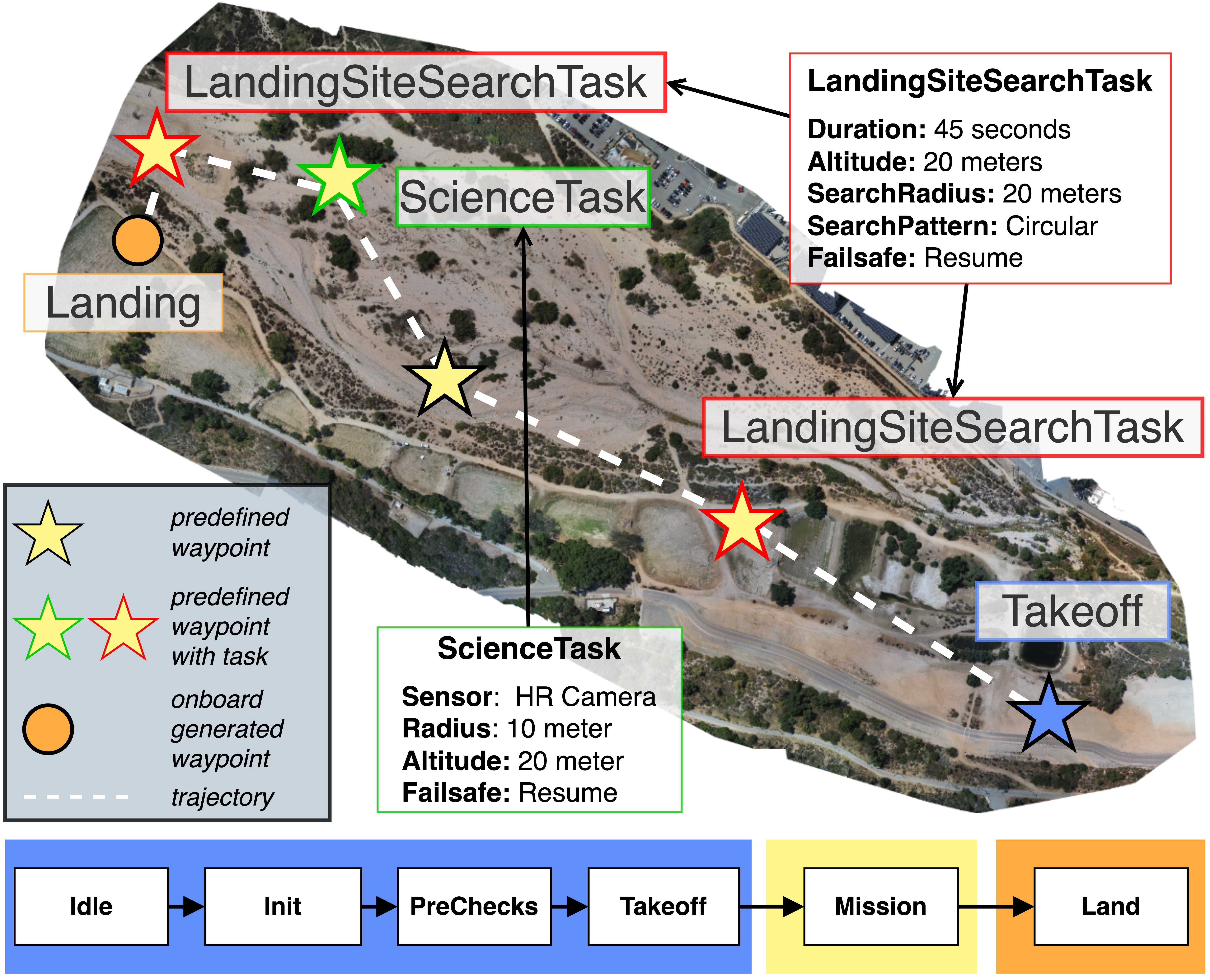}
    \caption{A notional MSH mission plan with structured task execution. The flight trajectory (dashed line) progresses through takeoff, mission navigation, and landing, with predefined waypoints (yellow) guiding execution. Tasks include LandingSiteSearchTasks (red) and  ScienceTasks (green). The FSM state sequence (bottom) is shown in simplified form: each transition reflects the successful execution of a state's BT (i.e., root node returns \texttt{Success}). This schematic omits explicit event triggers to emphasize overall mission flow.}
    \label{fig:ExampleMission}
\end{figure}

Aerial vehicles have revolutionized planetary exploration by enabling access to scientifically valuable but hazardous terrain beyond the reach of ground-based robots. NASA’s Ingenuity Mars helicopter demonstrated the feasibility of controlled flight on Mars, completing 72 successful flights despite the planet’s thin atmosphere \cite{balaram_ingenuity_2021}, \cite{golombek2022mars}. However, Ingenuity was a technology demonstrator, designed primarily to validate powered flight rather than autonomously execute complex scientific missions. The next-generation MSH aims to expand these capabilities, conducting long-range science missions across diverse Martian landscapes while autonomously navigating unstructured terrain and performing real-time scientific tasks. Figure~\ref{fig:ExampleMission} illustrates a representative MSH mission scenario, including takeoff, navigation, and landing phases, with operational and scientific tasks assigned at pre-defined waypoints.

Future aerial space missions require a higher level of autonomy to ensure safe, efficient, and adaptive operation in partially unknown and unpredictable Martian environments. Unlike Ingenuity, which followed human-planned trajectories over relatively flat terrain, MSH must traverse diverse landscapes, requiring adaptive risk management and real-time decision-making to balance mission objectives with vehicle safety. These requirements introduce several fundamental autonomy challenges that must be addressed for successful aerial exploration:

\begin{itemize}
    \item \textbf{Scalable and Modular Autonomy} – A structured control architecture is required for adaptive, reactive, and deterministic behavior execution. The system must support explainable decision-making while enabling efficient synthesis, analysis, and modification of autonomous behaviors to ensure mission flexibility and aerospace-grade reliability.
     \item \textbf{Computational Efficiency \& Real-Time Adaptation} – Limited onboard processing requires computationally efficient algorithms that enable real-time decision-making.
    \item \textbf{Adaptive Landing Site Detection} – MSH must autonomously search, evaluate, and prioritize landing sites during flight, ensuring safe emergency landings in hazardous regions. Unlike Ingenuity, which relied on pre-defined human-designated landing zones, MSH must continuously detect and maintain fallback landing sites to ensure safe operations over hazardous and partially unknown terrain, reducing the risk of mission loss.
    \item \textbf{Fault-Tolerant Decision-Making} – Deep-space autonomy must monitor system health (e.g., battery state) and trigger appropriate fail-safe behaviors. The system must autonomously reconfigure, deciding whether to continue, adjust flight parameters, or abort the mission based on real-time conditions.
    \item \textbf{Human-Guided Task Assignment} – Operators define high-level mission objectives using satellite imagery, but the system must dynamically adapt task execution based on real-time scientific discoveries and environmental conditions, ensuring mission success while maintaining vehicle safety.
\end{itemize}

To address these challenges, this paper presents a hybrid autonomy framework for deep-space aerial exploration, integrating FSMs and BTs. The FSM provides structured, deterministic state transitions, while BTs enable modular, reactive task execution, allowing the system to dynamically adapt to system and environmental conditions. This combination balances high-level mission structure with fine-grained task flexibility, ensuring robust decision-making under constrained computational resources.

The framework operates on an event-driven behavior adaptation mechanism, allowing autonomous responses to system health changes, environmental conditions, and mission task execution. By continuously monitoring vehicle state, battery levels, and onboard anomalies, the system triggers adaptive mission reconfiguration or fail-safe actions as needed. To support scalability and reusability, the architecture is modular, facilitating extensibility, structured testing, and automated verification while maintaining aerospace-grade reliability.

Additionally, the framework allows human-guided mission design, enabling operators to predefine mission objectives and conditional tasks based on satellite imagery while retaining the system’s ability to execute adaptive tasks based on real-time discoveries. Designed to be middleware-agnostic, it supports both ROS-based prototyping and space-grade middleware such as F-Prime, ensuring applicability across different robotic platforms and mission scenarios \cite{quigley2009ros}, \cite{Bocchino2022FPP}.

The primary contributions of this paper include:

\begin{itemize}
    \item A hybrid autonomy framework that integrates deterministic mission execution with reactive task adaptation, ensuring structured yet flexible autonomous control.
    \item A two-tier decision-making architecture, where the Autonomy module orchestrates mission execution, while the Healthguard continuously monitors system health and triggers adaptive responses.
    \item A middleware-agnostic implementation, enabling seamless integration with ROS-based development and space-certified middleware (F-Prime) for real-world deployment.
    \item A modular and scalable architecture, allowing for extensible behavior control, dynamic task execution, and robust failure handling across diverse mission scenarios.
    \item Extensive Monte Carlo simulations and real-world validation, confirming the system’s robustness, resilience to failures, and ability to autonomously respond to mission-critical conditions.
\end{itemize}

\section{Related Works} \label{Related_Works}

Several researchers have explored hybrid approaches integrating FSMs and BTs to leverage the strengths of both paradigms in autonomous robotic systems. FSMs are well-established for their deterministic simple design and structured transitions, making them particularly suitable for high-level mission phases such as takeoff, landing, and mission-mode switching. These characteristics are critical in domains where predictability and formal verification are paramount, such as space exploration and safety-critical robotics. Despite their advantages, FSMs face limitations when dealing with complex, dynamic, or concurrent tasks. As state complexity grows, FSMs become difficult to maintain, leading to state explosion and reduced scalability. To address these limitations, BTs have emerged as a powerful alternative due to their modularity, hierarchical structure, and support for concurrent task execution. Unlike FSMs, which rely on rigid transitions between states, BTs encapsulate behaviors in a structured manner that allows for flexible control flow, efficient failure handling, and reactive decision-making \cite{bojic2011extending}, \cite{ghzouli2023behavior}, \cite{iovino2024comparison}, \cite{Colledanchise_2018}, \cite{iovino2022programmingeffortrequiredgenerate}.

A key motivation for using FSMs at the high level while employing BTs within states is the need for structured decision-making alongside dynamic execution. FSMs ensure strict, deterministic transitions between well-defined mission phases, thereby enforcing mission-critical constraints and state-dependent operational safety. In contrast, BTs provide reactivity and modularity within each state, allowing the system to adapt to real-time conditions without modifying the global state structure. This hybridization reduces the complexity of state transition management while maintaining the benefits of adaptable task execution \cite{iovino2022survey}.

While BTs offer a modular approach to task execution, their recursive evaluation model inherently introduces hidden state dependencies and unnecessary computational overhead \cite{chen2018development}. Although backchaining BTs can mitigate some of these issues, it significantly increases mission design complexity—an approach that is currently impractical given the constraints and mission-critical nature of aerial space exploration. In contrast, FSMs provide an event-driven execution model with lower processing time per tick, making them more suitable for mission-level transitions where strict transition logic are necessary \cite{iovino2024comparison}.

Furthermore, hybrid architectures such as Aerostack2 have explored multi-agent aerial robotics frameworks that integrate BTs with FSMs. However, their reliance on ROS-based middleware, services, and actions introduces complexity unsuitable for space-constrained environments. Space-grade autonomy frameworks, such as those built on F-Prime, require lightweight, computationally efficient, and deterministic execution models. The hybrid FSM-BT approach aligns with these requirements by ensuring robust, event-driven high-level state management while allowing for adaptive and modular behavior execution at lower levels \cite{fernandezcortizas2024aerostack2softwareframeworkdeveloping}, \cite{zutell2022flexiblebehaviortreessearch}, \cite{mohta2018fast}, \cite{foehn2022agilicious}, \cite{wuthier2021productivemultitaskingindustrialrobots}.

A few studies have been published specifically for space applications focusing on task management and behavior control of robotic systems. For space exploration, NASA's Jet Propulsion Laboratory (JPL) developed frameworks like TRACE, which uses Business Process Model Notation (BPMN) and ROS actionlib to coordinate robotic behavior \cite{de2020event}. However, ROS limitations—such as inadequate real-time performance, lack of deterministic behavior, and insufficient fault tolerance—make TRACE unsuitable for space-grade applications. Another relevant framework, MEXEC (Multi-mission EXECutive), was initially developed for orbital spacecraft, such as the Europa Lander mission \cite{verma2017autonomous}. MEXEC excels in re-planning and task-net scheduling but is less suited for aerial vehicles that require rapid responses. The framework relies on an accurate model of its resources, such as battery voltage, which is problematic for multi-rotor vehicles with nonlinear and complex power consumption. Moreover, if the framework has to re-plan an action, it blocks subsequent adjustments for a predefined period. While this blocking is not critical for ground-based robots like CADRE (which can remain stationary while waiting), it limits the responsiveness of the framework, which is essential for aerial vehicles that need to react quickly in dynamic environments \cite{verma2017autonomous}, \cite{ingham2024onboard}.
    
While several mission control frameworks exist, none are specifically designed for autonomous aerial operations on planetary bodies like Mars. Existing solutions often lack the flexibility for complex missions or add unnecessary complexity. Many rely on ROS, making them incompatible with space systems that require a streamlined, middleware-agnostic approach to switch quickly to F-Prime.


\section{Autonomy Framework} \label{Implementation}

\begin{figure}[!t]
    \centering
    \includegraphics[width=0.9\columnwidth]{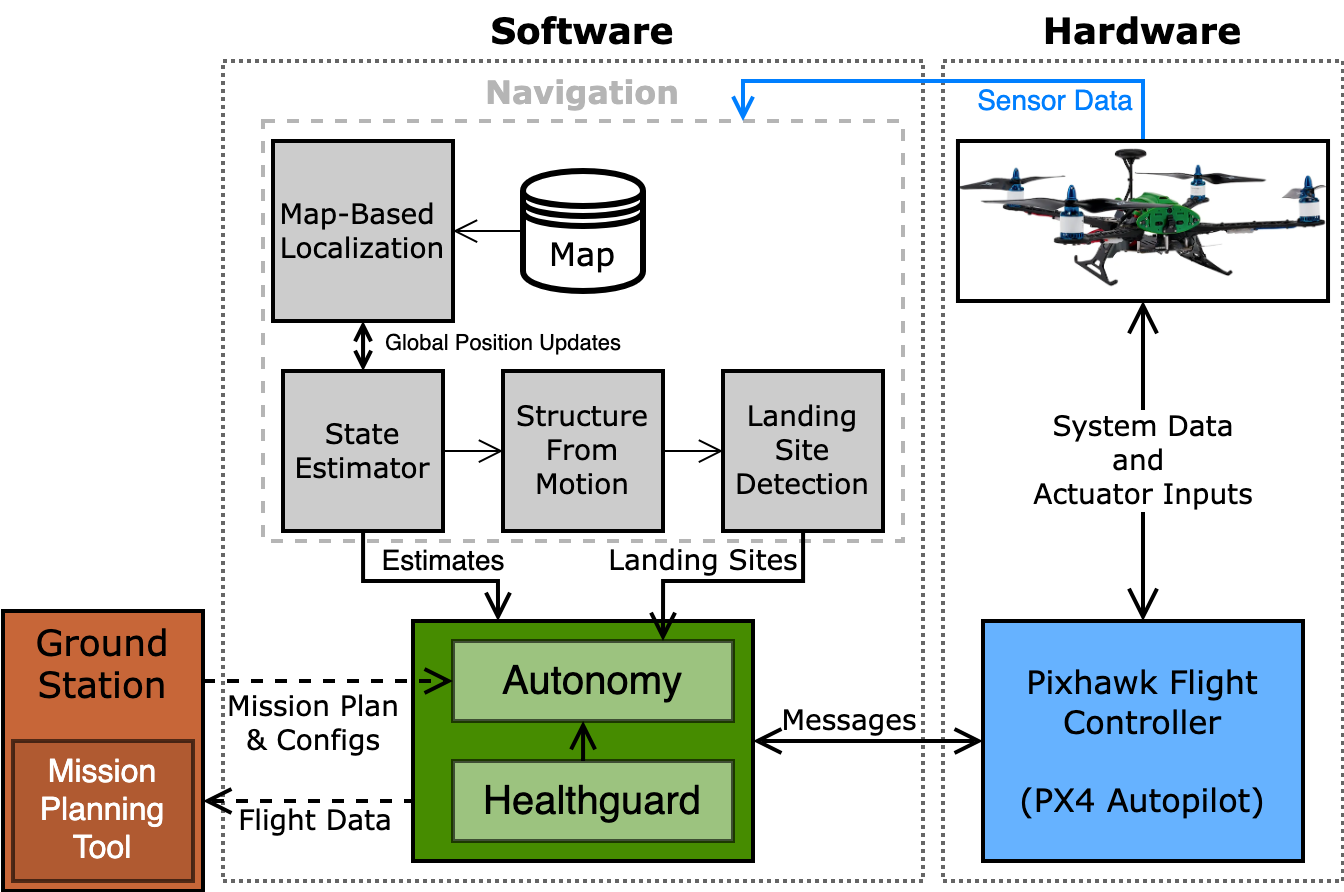}
    \caption{System architecture for autonomous mission execution. The Autonomy module orchestrates mission behavior, with the Healthguard monitoring system health and reporting anomalies. The Navigation stack provides real-time position updates using state estimation, map-based localization, structure from motion, and landing site detection. State estimates, including pose and velocity, are fed into the PX4 EKF2 for low-level control, while landing site management is handled by the Autonomy.}
    \label{fig:SystemOverview}
\end{figure}

Executing a complex mission autonomously requires a combination of different software and hardware elements, as can be seen in Figure \ref{fig:SystemOverview}. This subsection focuses on the software elements while the hardware side will be discussed in Chapter \ref{Integration_Testing}.
    
\subsection{System Architecture}

The mission autonomy framework, including the Autonomy and Healthguard modules, manages mission execution and ensures system safety. Prior to the operation, the Autonomy necessitates a mission plan generated with the mission planning tool on the ground station computer. This tool allows users to create specific science and operational tasks at predefined waypoints, instructing the Autonomy to execute tasks at designated locations, as illustrated in Figure \ref{fig:ExampleMission}. The FSM-BT-based Autonomy carries out the mission plan, guiding the helicopter's behavior. It sends high-level commands to the PX4 Autopilot on a commercial flight controller to manage navigation and task execution. Simultaneously, the Healthguard module consistently monitors crucial parameters such as battery levels, actuator status, and the confidence of the state estimator's estimations. In case of anomalies, the Healthguard notifies the Autonomy module, which then takes requisite actions to maintain the vehicle's safety and operational continuity. The decision-making process within the framework uses the outputs from other software modules in the stack.

The software stack includes xVIO, a visual-inertial odometry-based state estimator that provides pose estimation of the vehicle during flight \cite{delaune2020xviorangevisualinertialodometryframework}. To ensure accurate global navigation during long-range missions and account for drift, a map-based localization (MBL) algorithm is utilized. MBL corrects the vehicle's position in a global reference frame using pre-recorded maps from the Mars Reconnaissance Orbiter \cite{brockers2022board}. Furthermore, the system incorporates Structure from Motion and Landing Site Detection (LSD) modules, enabling real-time detection of safe landing sites using a monocular camera. These modules assess terrain roughness, slope, and available space to ensure a safe landing in case of an emergency or planned landing scenario \cite{proencca2022optimizing}, \cite{schoppmann2021multiresolutionelevationmappingsafe}, \cite{brockers2021autonomous}. The Autonomy uses the landing sites and pose estimates to coordinate the helicopter's navigation and behavior throughout the mission.

\subsection{Mission Autonomy}

\begin{figure}[!t]
    \centering
    \includegraphics[width=0.9\columnwidth]{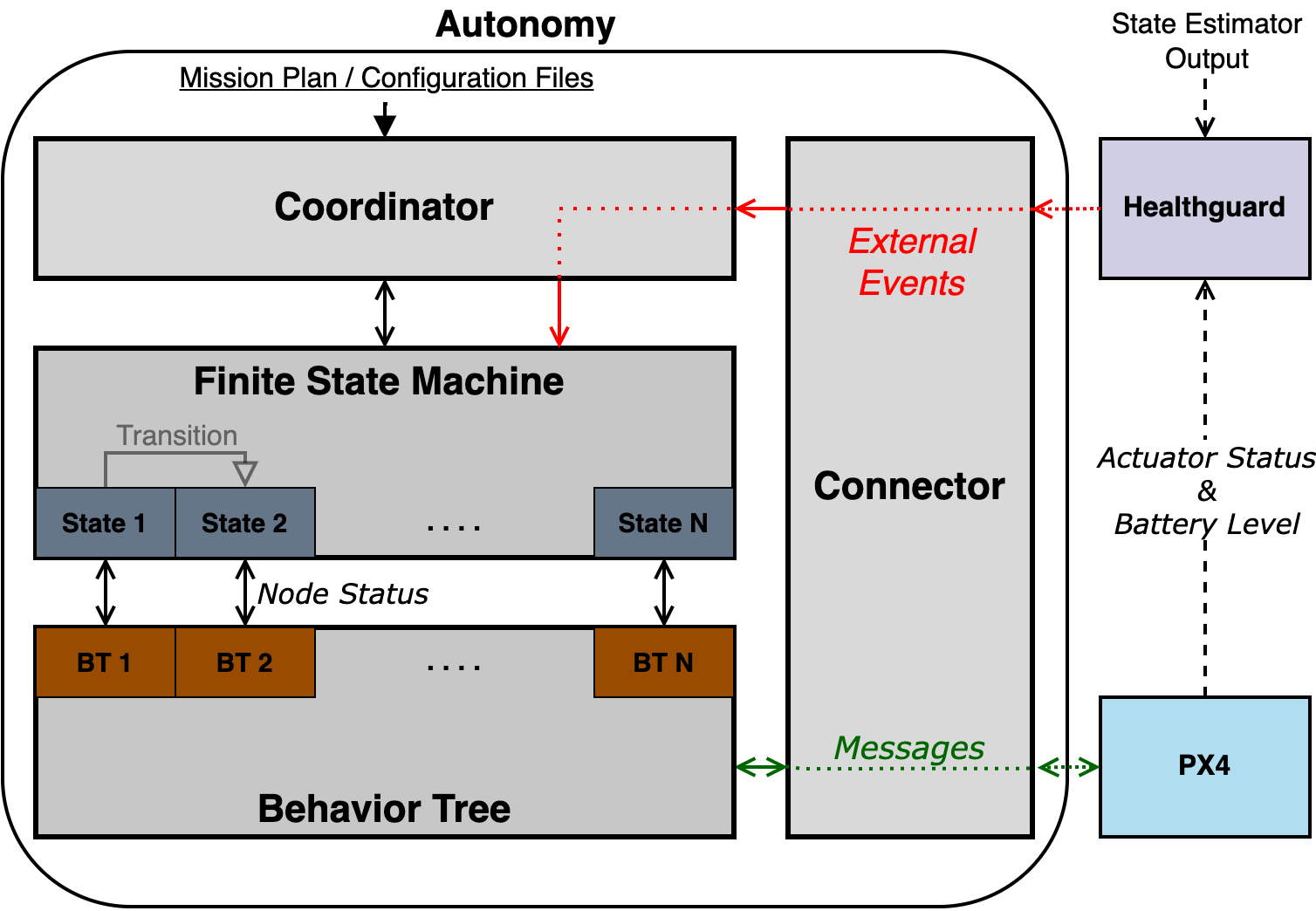}
    \caption{Mission autonomy framework with core software components and data flow. The Coordinator oversees execution, managing the FSM for deterministic mission phases and BTs for flexible, reactive task execution. The Healthguard monitors system health, detecting anomalies in actuator status, battery levels, and estimator confidence, and sends external events to the Coordinator, triggering adaptive responses. The Connector enables middleware-agnostic communication, facilitating data exchange between the autonomy system and PX4 for mission control. This architecture ensures robust autonomy, real-time adaptation, and compatibility with ROS and space-grade middleware like F-Prime.}
    \label{fig:AutonomyFW}
\end{figure}

The Autonomy is structured around modular components that provide flexible, adaptive control for autonomous aerial missions. The core elements of the framework illustrated in Figure \ref{fig:AutonomyFW} include the \textit{Coordinator}, which manages system initialization and execution, the \textit{Finite State Machine}, and \textit{Behavior Tree} for decision-making and task execution, as well as the \textit{Connector}, which serves as the interface for communication with external systems. Supporting modules like the ParameterServer, GlobalBlackboard, and MissionParser provide the system's configuration, mission planning, and data management capabilities.

The \textit{Coordinator} serves as the primary interface in the Autonomy, initializing core and support modules, managing asynchronous execution by running the FSM, and processing events. Events are forwarded to the FSM, ensuring that state transitions occur under specific conditions and priorities.

The \textit{Finite State Machine} manages mission states and transitions using a state transition table. Each state represents a mission phase — Idle, Init, Takeoff, Mission, Land, Terminate or EmergencyLand — and is implemented as a separate class derived from an abstract base state class, ensuring extensibility for new mission profiles or vehicle configurations. Transitions are triggered deterministically by predefined events, which may originate from behavior trees returned node status or external sources, such as the Healthguard. External events such as \texttt{BatteryLow} or \texttt{BatteryCritical} are emitted by the Healthguard when predefined thresholds are violated. Each state defines explicit transitions for a subset of such events. If an event occurs without a defined transition, it is treated as a self-transition to the current state, maintaining execution flow. The FSM is validated offline to ensure that all states are reachable and that from each state, a path to a final state exists. This guarantees deadlock freedom and ensures system stability and predictability.

Each state's logic is governed by its corresponding \textit{Behavior Tree}, which the FSM activates based on the mission phase. BTs dictate the robot's actions during each phase and are generated by the Behavior Tree Factory, which links registered nodes for that state. The FSM can execute, pause, abort, or reset BTs based on mission needs. BTs are asynchronously ticked, and the node status (Success, Failure, or Running) propagates to the FSM. This interaction between FSM and BT enables high-level deterministic control while allowing for reactive behavior at the task level, adapting dynamically to real-time conditions.

A key feature of the FSM-BT architecture is its event-driven nature, where events arise from internal and external sources. Internal events, such as Success or Failure, which represent the BT status and are required by all states, trigger state transitions. Additionally, more specific internal events, like "landing site search failed" after multiple retries, can be defined to handle particular mission conditions. External events, such as a Healthguard or other external software running in the stack, may report critical system parameters or other conditions that could cause a behavior change. The Healthguard simplifies BTs by reducing the need for repetitive condition nodes, which would otherwise add complexity. The \textit{Coordinator} processes all events and determines when a state transition is necessary, enabling the system to adapt dynamically without manual intervention—such as triggering an emergency landing due to critical battery levels.

The \textit{Behavior Tree} structure provides flexible control over low-level behaviors. Control nodes (Sequence, Fallback, Parallel) manage task execution order, while decorator nodes (retry, timeout, inverse) modify the behavior of their child nodes. Leaf nodes represent action and condition nodes that execute specific actions or check conditions. The modularity of BTs allows user-defined action and condition nodes to be reused across multiple states, enhancing flexibility. Nodes can be dynamically adjusted—for example, marking nodes as skipped during the next tick—allowing BTs to adapt in real-time without state transitions. Within BTs, fallback nodes provide a natural recovery mechanism by redirecting execution when leaf nodes fail. Additionally, BTs can directly trigger state transitions by notifying the state machine through predefined internal events, such as a task failure.

The \textit{Connector} decouples the autonomy system from specific middleware or hardware, providing a middleware-agnostic interface that facilitates communication with external systems like the PX4 flight controller, xVIO State Estimator, and Healthguard. This interface ensures seamless data exchange and real-time mission command processing. The Connector allows integration with different platforms, such as ROS, F-Prime, and PX4, without altering the core mission logic, offering flexibility across hardware architectures. In the current implementation, ROS 1, MAVROS, and PX4-specific commands and interactions are supported.

\begin{figure*}[!hbt]
    \centering
    \includegraphics[width=1.9\columnwidth]{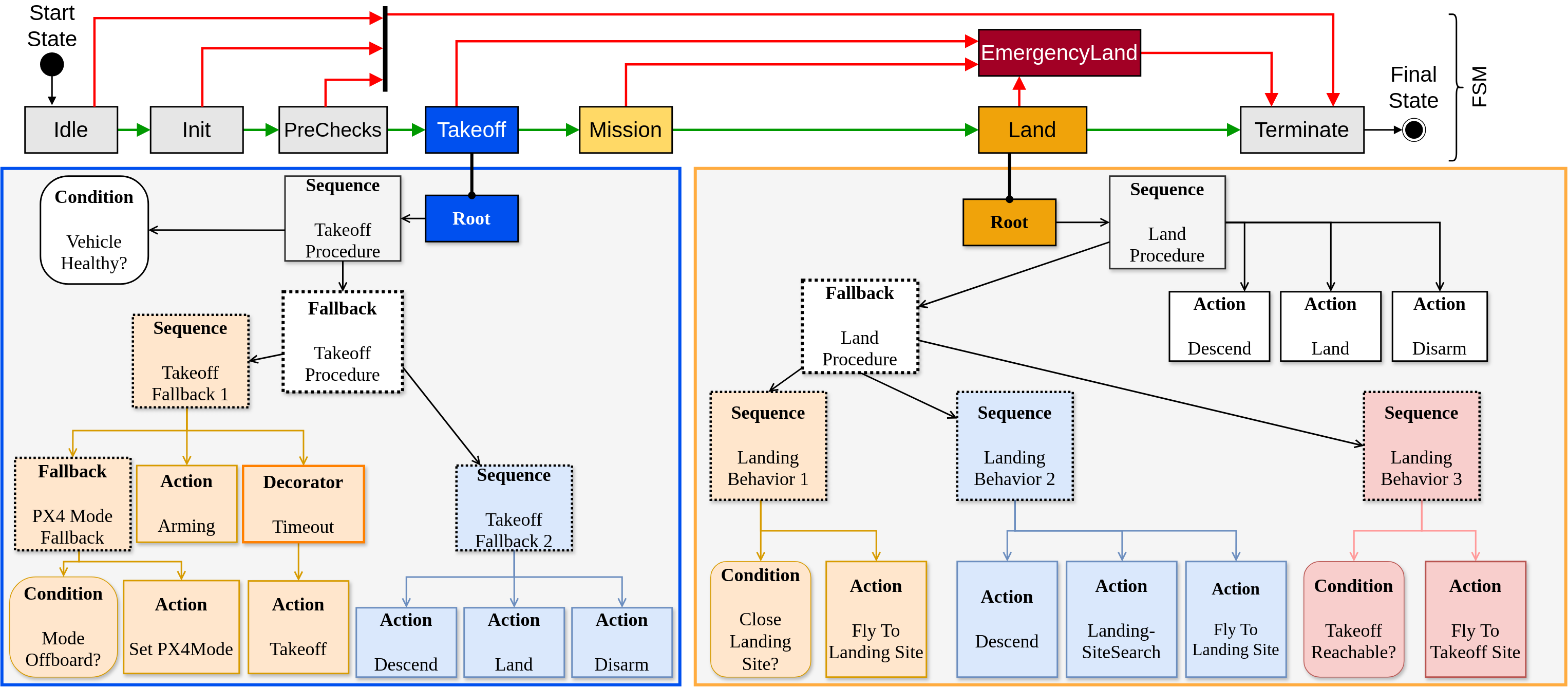}
    \caption{High-level FSM governing mission phases (top) and integrated BTs managing task execution within each phase (bottom). The FSM is shown in schematic form: green arrows represent transitions upon successful BT execution (i.e., root returns \texttt{Success}), while red arrows capture transitions triggered by failure or system-level events such as safety violations or health events. Multiple events may lead to the same transition; arrows summarize such cases rather than depicting each trigger individually. This abstraction emphasizes mission logic over full specification. Within each state, BTs enable reactive execution using control nodes (Sequence, Fallback), decorators (e.g., Timeout), and leaf nodes (Condition, Action). The left (blue) box shows takeoff procedures; the right box represents landing, including real-time landing site search and fallback strategies. The hybrid FSM-BT design combines predictable state transitions with dynamic, condition-driven task execution.}
    \label{fig:FSM_BT_Integration}
    \vspace{-4mm}
\end{figure*}

Figure \ref{fig:FSM_BT_Integration} illustrates the core FSM-BT control structure, showing the primary states: Idle, Init, PreChecks, Takeoff, Mission, Land, EmergencyLand, and Terminate. Additionally, it shows the behavior trees corresponding to the Takeoff and Land states. Before any action is taken, the BT first checks the vehicle's health status. If the check is passed, the Takeoff behavior tree will execute the nodes of the fallback sequence node (orange) on the left. The first task is setting the PX4 mode to Offboard and then arming the vehicle. Once the vehicle is armed, the actual Takeoff action is executed, controlling the vehicle to ascend using lateral position control and vertical velocity control. A timeout is applied to the takeoff action node, which is dynamically calculated based on the vehicle's velocity and distance to the takeoff waypoint. If any of these leaf nodes fail, the nodes of the fallback sequence node on the right will be executed. The Descend, Land, and Disarm actions are executed sequentially. If one of these actions also fails, the behavior tree will return a Failure status, causing a transition to the EmergencyLand state. Entering the Land state, after a successful mission execution (which involves completing the mission plan), the vehicle attempts to land at the closest identified landing site. A search pattern is flown at a lower altitude if no landing site is available to locate suitable landing sites. Once a landing site is found, the system targets the most confident site and proceeds to land. Again, if the central fallback sequence nodes are failing, a transition to the EmergencyLand is executed.

\section{Integration and Testing} \label{Integration_Testing}

During the development and evaluation phase of the high-level Autonomy software, extensive simulations and flight tests were conducted to refine, extend, and verify the framework's robustness and functionality. This section presents the experiments and final results achieved in this research.

The hardware setup for our experiments included the ModalAI Sentinel drone, which was equipped with advanced sensors for GPS-denied navigation and landing site detection. A Lightware Laser Range Finder (LRF) was integrated with a downward-tilted 45-degree RGB and stereo cameras. The core of the drone's system is the VOXL2 onboard computer, which communicates with the Pixhawk 6C Mini flight controller via a UART-to-USB connection. While the PX4 ran on the flight controller and the mission planning tool on the ground station, all presented software, including the mission autonomy framework, was executed on the VOXL2. Additional components included a telemetry antenna and a voltage measurement sensor, which provided real-time telemetry data to the ground station computer. The ground station was also used to upload flight plans and adjust flight parameters, facilitating the execution of different autonomous flight missions orchestrated by the Autonomy and ensuring safety during experimental flights.

\subsection{Monte-Carlo Simulations}

To systematically evaluate the robustness of the autonomy framework, two major Monte Carlo simulation series were conducted in Gazebo Garden with integrated PX4 firmware, ensuring close alignment with real hardware execution. The simulations tested both the Autonomy Module, responsible for orchestrating mission execution and reacting to system events, and the Healthguard Module, which continuously monitors critical parameters and generates discrete event triggers for adaptive decision-making. This structured approach allowed for an independent assessment of both mission control and system health monitoring.

\begin{figure}[!t]
    \centering
    \includegraphics[width=1.0\columnwidth]{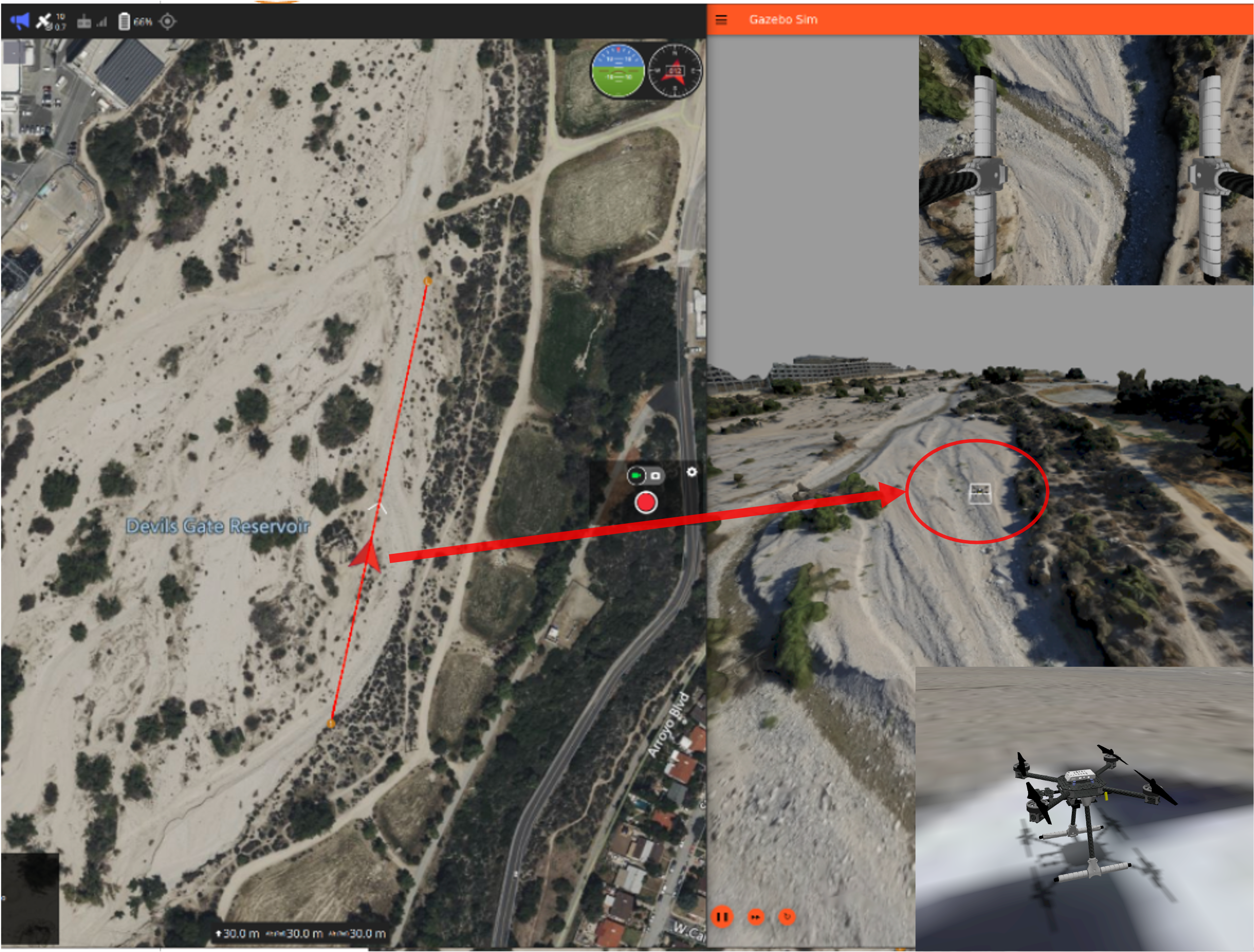}
    \caption{Simulation environment in Gazebo replicating real-world field conditions, integrating a high-fidelity drone model with onboard sensors identical to the real system. The terrain was generated from aerial mapping data and processed in Blender to create a high-resolution, georeferenced 3D environment, ensuring accurate terrain representation. While Gazebo does not provide fully photorealistic rendering, this setup enables realistic physics-based testing of the autonomy framework, allowing for meaningful validation of mission-critical behaviors, perception algorithms, and sensor-driven decision-making before real-world deployment.}
    \label{fig:Gazebo_Simulation}
    \vspace{-4mm}
\end{figure}

The simulation environment (Figure \ref{fig:Gazebo_Simulation}) was generated from real-world aerial mapping data processed in Blender, creating a high-resolution, georeferenced 3D terrain model. While Gazebo does not provide fully photorealistic rendering, it accurately replicated sensor inputs, flight dynamics, and terrain conditions, making the tests highly representative of field conditions.

The first Monte Carlo simulation series evaluated the Autonomy module's ability to react to system events, ensuring seamless coordination between behavior trees, finite state transitions, and navigation execution. Across 170 randomized trials, the system encountered events such as battery depletion, state estimation faults, and failure to detect safe landing sites, testing its ability to dynamically reconfigure mission execution while maintaining safety constraints. The autonomy consistently performed correct state transitions, successfully handling all predefined failure scenarios. The event distribution and corresponding state transitions are summarized in Table \ref{tab:simulations}.

The first Monte Carlo simulation series evaluated the Autonomy Module’s reaction to system events, verifying correct state transitions and dynamic mission adaptation. Across 170 trials, random health events were injected during a fixed mission plan to evaluate the system's response to various conditions, particularly those reported by the Healthguard. Events such as \textit{BatteryLow} triggered a landing, including a landing site search task. In more severe cases, like \textit{BatteryCritical} or \textit{StateEstimatorFailure} events, the system initiated an emergency landing with a rapid descent. The results consistently demonstrated the system's ability to correctly handle all predefined events and execute appropriate state transitions, ultimately reaching the final state, Terminate. Table \ref{tab:simulations} shows the distribution of health events and the FSM states in which they occurred. Furthermore, the Healthguard was tested in isolation to assess its robustness in detecting conditions such as critically low battery levels or estimation inconsistencies. Simulated input values were used to trigger Healthguard events, confirming its ability to reliably detect system anomalies.

    \begin{table}[b!]
        \caption{Distribution of 170 successful simulation results categorized by specific events randomly triggered across different states of the FSM.}
        \label{tab:simulations}
        \centering
        \begin{tabular}{|l|c|c|c|c|c|}
        \hline
        \textbf{Event}                  & \textbf{Init} & \textbf{Takeoff} & \textbf{Mission} & \textbf{Land} & \textbf{Tot.} \\ \hline
        \textit{StateEstimatorFailure} & 7  & 7  & 8  & 9  & 31 \\ \hline
        \textit{BatteryLow}       & 10 & 8  & 6  & 9  & 33 \\ \hline
        \textit{BatteryCritical}  & 5  & 8  & 5  & 7  & 25 \\ \hline
        \textit{EmergencyBattery} & 6  & 7  & 6  & 8  & 27 \\ \hline
        \textit{NoLandingSitesFound}     & 5  & 8  & 5  & 7  & 25 \\ \hline
        \textit{LandingSiteChecks}       & 7  & 9  & 7  & 6  & 29 \\ \hline
        \textbf{Total}                   & 40 & 47 & 37 & 46 & 170 \\ \hline
        \end{tabular}
    \end{table}

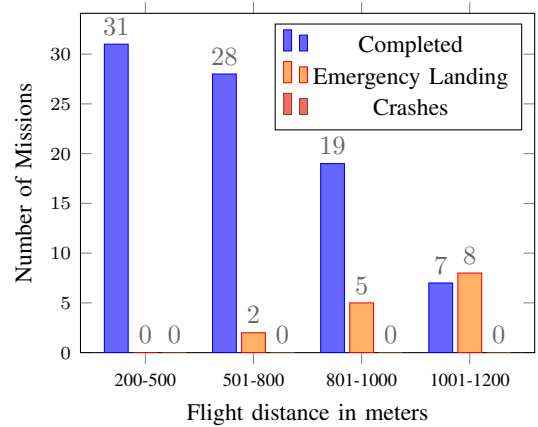
\begin{figure}[t!]
    \centering
    \begin{tikzpicture}
    \begin{axis}[
        ybar,
        bar width=9pt,
        ylabel={Number of Missions},
        xlabel={Flight distance in meters},
        symbolic x coords={200-500, 501-800, 801-1000, 1001-1200},
        xtick=data,
        enlarge x limits=0.2,
        legend style={at={(0.7,0.97)}, anchor=north, legend columns=1, font=\small},
        ymin=0,
        ytick distance=5,
        nodes near coords,
        nodes near coords align={vertical},
        width=3in,
        height=2.4in,
        tick label style={font=\scriptsize}, 
        label style={font=\small} 
    ]
    \addplot+[ybar, fill=blue, fill opacity=0.6, text=black] plot coordinates {(200-500, 31) (501-800, 28) (801-1000, 19) (1001-1200, 7)};
    \addplot+[ybar, fill=orange, fill opacity=0.6, text=black] plot coordinates {(200-500, 0) (501-800, 2) (801-1000, 5) (1001-1200, 8)}; 
    \addplot+[ybar, fill=red, fill opacity=0.6, text=black] plot coordinates {(200-500, 0) (501-800, 0) (801-1000, 0) (1001-1200, 0)}; 
    \legend{Completed, Emergency Landing, Crashes}
    \end{axis}
    \end{tikzpicture}
    \caption{Distribution of various mission outcomes across different distance categories. Each bar represents a specific range of distances, indicating the number of missions completed under nominal conditions (blue) and ended with an emergency landing (orange). No mission ended with a crash (red).}
    \label{fig:missions_bar_chart}
    \vspace{-4mm}
    \end{figure}

The second simulation series validated the Healthguard Module, ensuring accurate detection and timely reporting of system anomalies while assessing its impact on overall mission robustness. Across 100 Monte Carlo trials, various failure conditions, including battery degradation, sensor failures, and degraded state estimator confidence levels, were introduced to evaluate detection accuracy, event propagation latency, and differentiation between transient anomalies and mission-critical conditions. The Healthguard achieved over 98.5\% detection accuracy, reliably identifying faults, and enabling proactive mission adjustments. These trials also examined the interplay between mission planning, system health monitoring, and emergency handling, with randomized waypoint distributions (5–15) and flight distances up to 1.2 km. As expected, longer mission durations increased the frequency of \textit{BatteryLow} and \textit{BatteryCritical} events, leading to adaptive landing site searches, trajectory modifications, and failsafe activations. Figure \ref{fig:missions_bar_chart} illustrates the correlation between flight distance and emergency landings, confirming the autonomy framework’s ability to dynamically reconfigure mission execution while prioritizing vehicle safety.

Using high-fidelity simulations and controlled event injections, these tests confirmed that the autonomy framework reliably executes state transitions, enforces safety constraints, and dynamically adapts to real-time mission conditions. The ability to coordinate adaptive mission behavior with system health monitoring ensures robust, scalable, and resilient autonomous operations for deep-space aerial missions.

\subsection{Field Tests}

In addition to the simulations, the Autonomy was tested on hardware. As already mentioned in Section \ref{Integration_Testing}, a modified ModalAI Sentinel drone was used. Initial flights were performed using a tether, to ensure safety while verifying that the system behaves as expected. After verifying that the drone's behavior matches the simulation results, anomalies were simulated during flight. For that, each health event listed in Table \ref{tab:simulations} was sent manually to the drone. To verify consistency, five flights per health event were conducted. In all 30 flights, the drone responded adequately to every health event.

\begin{figure}[!t]
    \centering
    \includegraphics[width=\columnwidth]{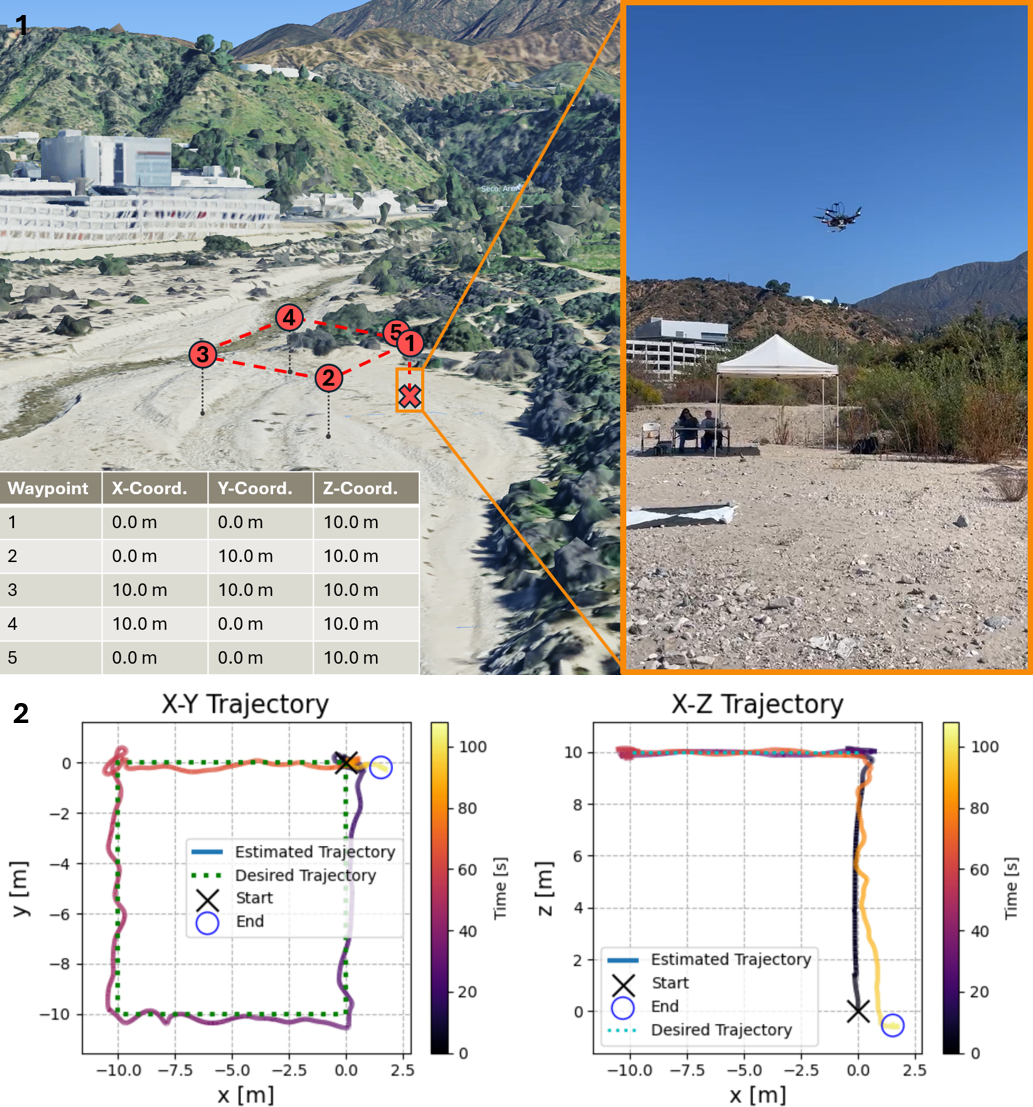}
    \caption{Autonomy and xVIO test flight in the Arroyo Seco, Pasadena, CA. Figure 1 provides an overview of the test site, including the predefined waypoints and the executed mission. The image shows also the drone in flight during testing. Figure 2 presents the recorded flight data, comparing the estimated trajectory against the desired trajectory in both the X-Y and X-Z planes, highlighting the accuracy of the state estimation and mission execution over time.}
    \label{fig:ArroyoTest}
    \vspace{-4mm}
\end{figure}

Following these tethered tests, the helicopter was put into a more realistic setting. Field tests were performed in the Arroyo Seco, a dessert-like terrain right outside of JPL. The goal of the field tests was to fly a mission fully autonomously. Multiple flights were conducted whereas one of them is shown in Figure \ref{fig:ArroyoTest}. The performed flight consisted of 10x10 m square at an altitude of 10 m. The drone successfully demonstrated autonomous takeoff, navigation and landing. For simplicity, only xVIO and Autonomy were used for that test. 

\subsection{Autonomy Performance and Reliability Evaluation}

Autonomy performance was evaluated based on key metrics, including control loop execution time, latency, memory consumption, and scalability.
The average control loop period on the VOXL2 was 2.2 ms, while latency between events and state transitions remained at 1.1 ms, ensuring prompt responses to external triggers. During flight tests, the VOXL2's CPU load averaged 6.5\%, reflecting efficient resource usage. Memory consumption remained stable at 24 MB throughout mission execution, with minimal increases during state transitions, even as behavior tree complexity grew.
Fault tolerance was tested through 270 Monte Carlo simulations and 30 test flights, where random health events—such as inconsistent state estimation or low battery—were triggered in different states. Each event was simulated at least three times per state, both in the Gazebo simulation environment and on the Sentinel drone. The autonomy system successfully handled all failures, transitioning to predefined fail-safe states. Additionally, random internal (e.g., BT Success, Failure) and external health events (e.g., low battery) were introduced to validate state transitions and fail-safe responses. The framework exhibited 100\% correctness in state transitions and maintained stability when encountering undefined events, with no crashes or incorrect transitions. The system consistently achieved 100\% mission completion, responding reliably to all events, including those that triggered failsafe behaviors, thus demonstrating its ability to execute long-range autonomous missions under diverse conditions.

\section{Conclusion and Future Work} \label{Conclusion}

Initially developed for the Mars Science Helicopter mission concept, the hybrid autonomy framework has proven its robustness and adaptability in various real-world applications, including a search and rescue drone project. The framework exhibited reliable performance across 400 development flights and competition runs, with no software failures or unexpected behavior, highlighting its resilience and potential for broader robotic and aerospace applications.

This research presents a flexible and middleware-agnostic framework that combines FSM and BT policies to enable sophisticated yet deterministic decision-making and high-level behavior control. Autonomy can execute reactive tasks while dynamically responding to real-time conditions, including handling emergencies such as rapid landings. A key strength of the framework is its Connector, which facilitates modular communication with diverse algorithms and supports seamless integration with various middleware and hardware platforms, making the system highly extensible across different robotic applications. The deterministic decision-making architecture of the FSM, combined with the reactivity and flexibility of the behavior trees, ensures robust mission execution with minimal complexity, allowing for precise control over the system's behavior while remaining adaptable to unforeseen circumstances.

This work lays the foundation for future autonomous systems in extraterrestrial and other environments, offering a versatile and scalable software architecture that can be adapted to various mission requirements and behavior patterns.

Future directions include dynamically triggering tasks based on real-time inputs, integrating auxiliary algorithms (e.g., scientific classifiers) to optimize mission strategy, and improving the behavior tree framework with graphical tools for easier configuration. Expanding to multi-agent capabilities will enable complex, cooperative missions that require coordinated decision-making among multiple autonomous agents. Additionally, implementing robust checks will ensure that the FSM and BT configurations are designed to always terminate in a valid final state, further improving system predictability and stability.

\section*{ACKNOWLEDGMENT}

This research was carried out at the Jet Propulsion Laboratory, California Institute of Technology, and was sponsored by the JPL Visiting Student Research Program (JVSRP) and the National Aeronautics and Space Administration (80NM0018D0004). 
All references to a future NASA Mars Science Helicopter concept are pre-decisional information and for planning and discussion purposes only.

\bibliographystyle{unsrt}
\bibliography{bibliography}

\end{document}